\documentclass[11pt]{article}
\usepackage{acl}
\usepackage{times}
\usepackage{latexsym}
\usepackage{amsmath}
\usepackage{amssymb}
\usepackage{booktabs}
\usepackage{graphicx}
\usepackage{multirow}
\usepackage{xcolor}
\usepackage{tikz}
\usepackage{pgfplots}
\pgfplotsset{compat=1.18}
\usetikzlibrary{shapes.geometric, arrows.meta, positioning, calc}

\title{Cross-Context Verification: Hierarchical Detection of\\Benchmark Contamination through Session-Isolated Analysis}

\author{Song Tae-Eun \\
  Daejeon Jungang Cheonggua Co., Ltd. \\
  \texttt{higheun@gmail.com}}

\begin{document}
\maketitle

\begin{abstract}
LLM coding benchmarks face a credibility crisis: widespread solution leakage and test quality issues undermine SWE-bench Verified, while existing detection methods---paraphrase consistency, n-gram overlap, perplexity analysis---never directly observe whether a model \emph{reasons} or \emph{recalls}. Meanwhile, simply repeating verification degrades accuracy: multi-turn review generates false positives faster than it discovers true errors \cite{song2026dccr}, suggesting that structural approaches are needed.

We introduce \textbf{Cross-Context Verification (CCV)}, a black-box method that solves the same benchmark problem in $N$ independent sessions and measures solution diversity, combined with the \textbf{Hierarchical Cross-Context Architecture (HCCA)}, a multi-agent analysis framework that prevents confirmation bias through intentional information restriction across specialized analytical roles.

On 9 SWE-bench Verified problems (45 trials, Claude Opus 4.6, temperature 0), CCV achieves \textbf{perfect separation} between contaminated and genuine reasoning (Mann-Whitney $U=0$, $p \approx 0.012$, $r = 1.0$). Key findings: (1)~contamination is binary---models either recall perfectly or not at all; (2)~reasoning absence is a perfect discriminator; (3)~33\% of prior contamination labels are false positives; (4)~HCCA's independent analysis structure discovers contamination-flaw composite cases that single-analyst approaches miss. A pilot experiment extending HCCA to multi-stage verification (Worker$\to$Verifier$\to$Director) yields a negative result---100\% sycophantic confirmation---providing further evidence that \emph{information restriction}, not structural complexity, is the key mechanism. We release all code and data.\footnote{Code and data: \url{[anonymized for review]}}
\end{abstract}

\section{Introduction}

SWE-bench Verified \cite{jimenez2024swebench} has become the de facto standard for evaluating LLM coding ability. However, its reliability is increasingly questioned. \citet{swebenchplus} found 32.67\% of successful patches involved solution leakage (solutions directly present in issue text or comments) through manual screening, while \citet{swebenchillusion} showed models recall file paths from training data up to 76\% of the time versus up to 53\% for external files. OpenAI's manual audit of 138 o3 failures revealed 59.4\% were caused by test flaws, not model limitations, leading them to recommend discontinuing SWE-bench Verified reporting.

Existing contamination detection methods have a shared limitation: they analyze \emph{artifacts} (text overlap, paraphrase consistency, perplexity) rather than directly observing the \emph{generation process}. CAP \cite{cap2024} tests paraphrase consistency but depends on perturbation quality. SWE-bench+ \cite{swebenchplus} manually screens issue text for solution leakage but never observes code generation. Perplexity-based methods \cite{deconIEP} require log-probability access, excluding most API-only models.

\paragraph{Key insight.} If a model has memorized a benchmark solution, it will reproduce the same answer regardless of session context. If it genuinely reasons, natural non-determinism---even at temperature 0---will produce varied solutions across independent sessions. This insight, inspired by prior work showing that independent-session review outperforms same-session review by 4.0 F1 points (16.3\% relative) \cite{ccr2026}, leads to a simple but powerful test: solve the same problem $N$ times in isolated sessions and measure diversity.

\paragraph{From repetition to structure.} A natural follow-up is whether repeated verification improves detection. Dynamic Cross-Context Review \cite{song2026dccr} tested this directly: multi-turn review with the same reviewer role \emph{degrades} performance (F1 drops from 0.376 to 0.303, $p < 0.001$), as additional rounds generate false positives faster than they discover true errors. This motivates our Hierarchical Cross-Context Architecture (HCCA): rather than repeating the same analysis, HCCA divides the task across specialized roles under strict information constraints (\S\ref{sec:hcca}). However, structural division alone is insufficient: a pilot experiment shows that when downstream agents can see upstream conclusions, sycophantic confirmation eliminates any filtering benefit (\S\ref{sec:pilot}), confirming that information restriction is the essential ingredient.

\paragraph{Contributions.}
\begin{enumerate}
\itemsep0em
\item \textbf{CCV method}: Session-isolated solution diversity as a black-box contamination and test flaw detector---the first to directly observe whether models reason or recall during code generation.
\item \textbf{HCCA framework}: A hierarchical multi-agent analysis architecture that prevents confirmation bias through intentional information restriction---extending CCR's context separation from the generation level to the analysis level.
\item \textbf{Empirical validation}: Perfect separation ($U=0$, $p \approx 0.012$, $r=1.0$) on 9 SWE-bench Verified problems, with HCCA's independent analysis structure discovering contamination-flaw composites that single-analyst approaches would miss.
\item \textbf{Information restriction as mechanism}: Converging evidence from three experiments---CCR, D-CCR, and an HCCA pilot---that session isolation, not structural complexity or repetition, is the necessary condition for effective verification.
\item \textbf{Practical tools}: Open-source contamination scoring formula, reasoning classifier (100\% accuracy on our data), and automated pipeline.
\end{enumerate}

\section{Related Work}

\paragraph{Benchmark contamination detection.}
Multiple approaches target contamination in LLM benchmarks. SWE-bench+ \cite{swebenchplus} manually screens issue text and comments for solution leakage, finding 32.67\% of successful patches involved direct solution presence. SWE-Bench Illusion \cite{swebenchillusion} diagnoses memorization versus reasoning through file path familiarity tests. CAP \cite{cap2024} measures consistency under paraphrase perturbation. Perplexity-based methods like DeconIEP \cite{deconIEP} detect memorized sequences but require log-probability access. \textbf{All existing methods share a gap}: none observe the generation process itself---whether the model analyzes, hypothesizes, and solves versus immediately producing a memorized patch. Table~\ref{tab:comparison} in \S\ref{sec:discussion} provides a detailed comparison.

\paragraph{SWE-bench test quality.}
OpenAI's audit of 138 o3 failures found 59.4\% attributable to test issues: restrictive tests (specific implementation enforcement), unspecified functionality testing, and environment dependencies.\footnote{OpenAI, ``SWE-bench Verified: Analysis of o3 Results,'' published via official blog, 2025.} SWE-bench Issue \#465\footnote{\url{https://github.com/princeton-nlp/SWE-bench/issues/465}} revealed git history leakage enabling models to access gold patches. These findings motivated SWE-bench Pro, Live, and rebench as alternatives.

\paragraph{Multi-agent verification.}
Multi-Agent Debate \cite{du2024debate} uses information-sharing discussion to improve accuracy. MetaGPT \cite{metagpt2024} assigns role-based collaboration. Self-Consistency \cite{wang2023selfconsistency} samples multiple solutions for majority voting. CCV differs fundamentally: it uses \emph{intentional information restriction} (session isolation) rather than information sharing, and measures \emph{process diversity} rather than answer agreement.

\paragraph{Multi-turn verification.}
\citet{song2026dccr} tested whether extending Cross-Context Review to multiple rounds (Dynamic CCR) improves verification. Single-pass CCR (F1 = 0.376) significantly outperformed all multi-turn variants (F1 = 0.263--0.303, $p < 0.001$), with \emph{false positive pressure} and \emph{Review Target Drift} as primary degradation mechanisms. The optimal number of review rounds is one---independent parallel reviews (ensemble) outperform sequential iteration. This negative result motivates HCCA's structural approach: dividing the analytical task across different roles rather than repeating the same role.

\section{Method: Cross-Context Verification}

\subsection{Session Isolation}

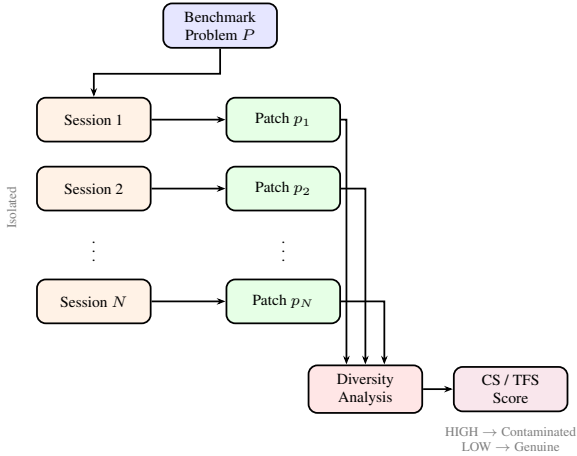
\begin{figure}[t]
\centering
\resizebox{\columnwidth}{!}{%
\begin{tikzpicture}[
    box/.style={draw, rounded corners, minimum width=1.8cm, minimum height=0.7cm, font=\scriptsize, align=center, thick},
    arrow/.style={-{Stealth[length=4pt]}, thick},
    label/.style={font=\tiny, text=gray}
]
\node[box, fill=blue!10] (prob) at (3.2,0) {Benchmark\\Problem $P$};
\node[box, fill=orange!10] (s1) at (1.2,-1.5) {Session 1};
\node[box, fill=orange!10] (s2) at (1.2,-2.6) {Session 2};
\node[font=\scriptsize] (dots) at (1.2,-3.5) {$\vdots$};
\node[box, fill=orange!10] (sn) at (1.2,-4.4) {Session $N$};
\node[label, rotate=90] (iso) at (-0.1,-2.9) {Isolated};
\node[box, fill=green!10] (p1) at (4.2,-1.5) {Patch $p_1$};
\node[box, fill=green!10] (p2) at (4.2,-2.6) {Patch $p_2$};
\node[font=\scriptsize] (pdots) at (4.2,-3.5) {$\vdots$};
\node[box, fill=green!10] (pn) at (4.2,-4.4) {Patch $p_N$};
\node[box, fill=red!10] (div) at (5.5,-5.8) {Diversity\\Analysis};
\node[box, fill=purple!10] (score) at (7.8,-5.8) {CS / TFS\\Score};
\draw[arrow] (prob.south) -- ++(0,-0.4) -| (s1.north);
\draw[arrow] (s1) -- (p1);
\draw[arrow] (s2) -- (p2);
\draw[arrow] (sn) -- (pn);
\draw[arrow] (p1.east) -| ([xshift=-0.3cm]div.north);
\draw[arrow] (p2.east) -| (div.north);
\draw[arrow] (pn.east) -| ([xshift=0.3cm]div.north);
\draw[arrow] (div) -- (score);
\node[label, below=0.1cm of score, align=center] {HIGH $\to$ Contaminated\\LOW $\to$ Genuine};
\end{tikzpicture}%
}
\caption{CCV pipeline. A benchmark problem is solved $N$ times in session-isolated trials. Solution diversity and gold proximity are analyzed to produce contamination and test flaw scores.}
\label{fig:pipeline}
\end{figure}

CCV operates on a simple principle: solve the same problem $N$ times in independent sessions, then analyze solution diversity (Figure~\ref{fig:pipeline}).

\paragraph{Session isolation.} Each trial uses: (1)~a fresh API conversation with no prior history, (2)~a clean repository clone at the problem's base commit, (3)~future git history purged (branches, tags, reflog), and (4)~identical prompts containing only the issue text (no hints). This ensures each trial is an independent observation of the model's behavior.

\paragraph{Core assumption.} Memorized recall is deterministic: a model that has seen the gold patch during training will reproduce it consistently. Genuine reasoning is non-deterministic: complex problems admit multiple valid approaches, and natural variation in LLM generation produces diverse solutions even at temperature 0. This non-determinism at temperature 0 arises from hardware and infrastructure-level factors in LLM inference, such as non-deterministic compute scheduling and co-mingled data in input buffers \cite{atil2024llmnondet}. Our control group confirms this empirically: pairwise similarity among genuine-reasoning problems ranges from 0.30 to 0.61 at temperature 0, derived from the per-pair similarity values underlying the diversity scores in Table~\ref{tab:full_trials} (Appendix~\ref{app:trials}).

\subsection{Diversity Metrics}

\paragraph{Code diversity.} For $N$ trials producing patches $p_1, \ldots, p_N$, we compute pairwise similarity using AST structural similarity (weight 0.4), line-level BLEU (0.3), and normalized edit distance (0.3):
\begin{equation}
\text{diversity} = 1 - \frac{1}{\binom{N}{2}} \sum_{i<j} \text{sim}(p_i, p_j)
\end{equation}

\paragraph{Gold proximity.} Mean and standard deviation of line-level BLEU between each trial patch and the gold patch:
$\bar{g} = \frac{1}{N}\sum_i \text{BLEU}(p_i, p_\text{gold})$, $\sigma_g = \text{std}(\cdot)$.

\subsection{Scoring Functions}

\paragraph{Contamination score.}
\begin{equation}
\text{CS} = 0.5\bar{g} + 0.3(1 - \text{diversity}) + 0.2(1 - \sigma_g)
\label{eq:contamination}
\end{equation}
Thresholds: HIGH $\geq 0.8$, MEDIUM $\geq 0.6$, LOW $< 0.6$.

The weights reflect the relative diagnostic value of each signal. Gold proximity receives the highest weight (0.5) as direct evidence of memorization---a model producing the gold patch verbatim is the strongest contamination indicator. Code diversity (0.3) captures whether the model explores the solution space, which is informative even when gold proximity is moderate. Gold proximity variance (0.2) detects consistency of recall; memorized solutions show zero variance across trials. These weights were set a priori based on the diagnostic reasoning above, not tuned on experimental data. Appendix~\ref{app:scoring} provides the full derivation for all scoring formulae.

\paragraph{Test flaw score.}
\begin{equation}
\text{TFS} = 0.4(1 - \text{FER}) + 0.4 \cdot \text{DBF} + 0.2 \cdot \text{diversity}
\label{eq:testflaw}
\end{equation}
where FER is the functional equivalence ratio (largest cluster of test-identical solutions / $N$) and DBF is the ratio of plausibly correct but test-failing solutions.

\paragraph{Correction for low-diversity test flaws.} When a memorized non-gold solution converges (diversity $< 0.05$) with low gold proximity ($\bar{g} < 0.5$), Eq.~\ref{eq:testflaw} underestimates test flaws because the diversity term vanishes. We introduce a \emph{memory-of-non-gold indicator}: if diversity $< 0.05 \wedge \bar{g} < 0.5$, we replace the diversity term with $0.2 \times (1 - \bar{g})$, yielding a corrected score that better reflects the test flaw evidence. In our data, this corrects \texttt{astropy-7606} from 0.400 to 0.762, aligning the quantitative score with qualitative evidence (gold patch fails in Verified).

\subsection{Reasoning Classifier}

We observe that contaminated models skip analysis and immediately output diffs, while genuinely reasoning models begin with problem analysis. Our classifier examines the first 100 tokens:
\begin{itemize}
\itemsep0em
\item \texttt{NO\_REASONING}: Output begins with \texttt{```diff} or patch content, total tokens $< 300$
\item \texttt{FULL\_REASONING}: Output begins with analytical phrases (``Looking at'', ``The issue is'', ``Let me analyze'')
\end{itemize}
This simple heuristic achieves 100\% accuracy on our 45 trials (15 contaminated = NO, 30 genuine = FULL).

We acknowledge a circularity concern: the classifier is evaluated on the same dataset used for contamination detection. However, the reasoning signal is \emph{behaviorally independent} of the diversity metrics---it captures \emph{how} the model begins its output (analytical preamble vs.\ immediate patch), not the patch content itself. A model could theoretically produce diverse patches while skipping reasoning, or identical patches with elaborate analysis; in practice, these signals are perfectly correlated in our data but measure different aspects of model behavior. Validation on held-out problems is needed to confirm generalizability (\S\ref{sec:limitations}).

\section{Hierarchical Cross-Context Architecture}
\label{sec:hcca}

If CCV detects contamination through solution diversity, a natural question is how to analyze the results without introducing analyst bias. Simply repeating the same analysis does not help: \citet{song2026dccr} showed that multi-turn review with a single role degrades performance as false positive pressure overwhelms marginal gains. HCCA takes a different approach---\emph{structural division of labor} across specialized roles with strict information constraints.

\subsection{Architecture}

HCCA organizes analysis into four layers with unidirectional information flow:

\begin{itemize}
\itemsep0em
\item \textbf{Layer 0} (Designer): Creates the experiment protocol, assigns problem groups to analysts, and defines information boundaries. Never touches experimental data.
\item \textbf{Layer 1} (Executor): Runs \texttt{solve\_in\_isolation()} to generate raw trial data---patches, metrics, and logs for each trial.
\item \textbf{Layer 2} (Analysts): Three independent analysts, each assigned one problem group (contamination suspects, test flaws, control). Each sees only Layer~1 data---never each other's analyses or conclusions.
\item \textbf{Layer 3} (Integrator): Combines Layer~2 results, performs cross-validation, resolves discrepancies, and computes final statistics.
\end{itemize}

Information flows strictly upward: L1 $\to$ L2 $\to$ L3, with no lateral communication at Layer~2. This prevents the confirmation bias that arises in shared-context systems \cite{du2024debate} and the anchoring effect that degrades multi-turn review \cite{song2026dccr}.

\subsection{Distinction from Multi-Agent Frameworks}

Conventional multi-agent systems---CrewAI, AutoGen, MetaGPT \cite{metagpt2024}---maximize information sharing between agents to improve coordination. HCCA deliberately restricts it:

\begin{itemize}
\itemsep0em
\item Layer~2 analysts see raw data but not each other's interpretations
\item The Integrator sees all Layer~2 outputs but performed no problem-level analysis
\item The Designer sets methodology but never sees results
\end{itemize}

This mirrors CCV's core insight applied at the analysis level: where CCV separates the model's \emph{solution contexts} to detect contamination, HCCA separates researchers' \emph{analytical contexts} to prevent confirmation bias. The principle is the same---intentional information restriction produces more reliable outcomes than unconstrained information sharing.

\subsection{Empirical Evidence}

Our experiment provides direct evidence that HCCA's structure matters. The Layer~2 analyst for Group~B (test flaws) classified \texttt{astropy-7606} as a pure test flaw based on external evidence of gold patch failure. The independent Layer~2 analyst for Group~C, performing cross-validation \emph{without seeing Group~B's analysis}, independently computed a contamination score of 0.641 and flagged it as a contamination-flaw \emph{composite}---a nuance that became a key finding (\S\ref{sec:testflaw}). This discovery emerged precisely because analysts operated with different analytical frames. A single analyst running all groups sequentially would likely have anchored on the first interpretation. Cross-validation showed 100\% numerical agreement between all independent analysts, confirming both reliability and the absence of analytical contamination across layers.

\subsection{Pilot Experiment: When Information Restriction Breaks}
\label{sec:pilot}

HCCA's CCV analysis succeeds because Layer~2 analysts never see each other's conclusions. But what happens when a multi-stage pipeline \emph{does} share conclusions between stages? We tested this with a Worker$\to$Verifier$\to$Director pipeline applied to CCR code review findings.

\paragraph{Setup.} We selected 3 artifacts from the CCR dataset (C1, D1, S1) and ran each through three conditions: (1)~\textbf{CCR-1}: single-pass review (baseline, F1 = 0.329 on these artifacts); (2)~\textbf{CCR-WV}: Worker generates findings, Verifier sees the findings and judges each as CONFIRMED or REJECTED; (3)~\textbf{CCR-WVD}: same as CCR-WV, plus a Director who reviews Verifier-confirmed findings and makes final ACCEPT/REJECT decisions. All stages used Claude Opus 4.6 at temperature 0.3.

\paragraph{Results.} The Verifier confirmed 100\% of Worker findings across all 3 artifacts (15/15 findings, all rated ``CONFIRMED'' with HIGH or MEDIUM confidence). The Director subsequently accepted 100\% of Verifier-confirmed findings (15/15). No filtering occurred at any stage: CCR-WV = CCR-WVD = CCR-1 (F1 = 0.329, identical).

\paragraph{Diagnosis.} The Verifier exhibited systematic sycophancy: when presented with a Worker's finding and its reasoning, the Verifier produced detailed, confident justifications for agreement in every case---even for findings that were borderline or arguable. This mirrors the anchoring effect identified in D-CCR \cite{song2026dccr}, where exposure to prior conclusions biases subsequent judgment. The critical difference from HCCA's CCV analysis is \emph{information flow}: in HCCA, Layer~2 analysts see raw data but never each other's interpretations; in this pilot, the Verifier saw the Worker's \emph{conclusions}, breaking the information restriction that makes independent judgment possible.

\paragraph{Implication.} Structural division of labor (Worker/Verifier/Director) without information restriction produces zero additional benefit. The result is consistent across three studies: CCR shows information restriction works; D-CCR shows repetition without new information fails; this pilot shows structural roles with shared conclusions also fail. The common factor is whether each stage makes judgments \emph{without seeing prior conclusions}.

\section{Experimental Setup}

\paragraph{Problems.} We select 9 SWE-bench Verified problems in 3 groups, drawn from 5 repositories to avoid repository bias.\footnote{We excluded 9 problems from the initial candidate pool due to duplicate repositories (\texttt{requests}: 4 problems), known environmental issues (\texttt{django-10914/10924}), or ambiguous contamination signals. See Appendix~\ref{app:selection} for the full exclusion list.} Selection criteria: Group~A required documented evidence of verbatim reproduction by at least one frontier model; Group~B required external evidence of test-side issues (gold patch failure, flaky tests, unspecified behavior); Group~C required no contamination or test flaw evidence and sub-50\% solve rates across public leaderboards.
\begin{itemize}
\itemsep0em
\item \textbf{Group A} (contamination suspected, 3): \texttt{django-11451} (GPT-5.2 verbatim reproduction reported), \texttt{django-11099} (Gemini 3 Flash reproduction), \texttt{astropy-13236} (Claude Opus 4.5 inline comment citation)
\item \textbf{Group B} (test flaws, 3): \texttt{astropy-7606} (gold patch fails in Verified), \texttt{matplotlib-20488} (flaky test, SSL issues), \texttt{django-10097} (missing template paths)
\item \textbf{Group C} (control, 3): \texttt{sklearn-14894}, \texttt{pytest-7571}, \texttt{xarray-3151} --- selected for no prior contamination evidence and sub-50\% solve rates
\end{itemize}

\paragraph{Model \& configuration.} Claude Opus 4.6 (\texttt{claude-opus-4-6}), temperature 0.0, 5 trials per problem, 45 total trials. Temperature 0 maximizes reproducibility; diversity arises from session isolation and hardware-level non-determinism (\S3.1).

\paragraph{Execution modes.} \emph{Lite mode}: issue text only, no repository clone---tests diversity metrics. \emph{Full mode}: Docker container (\texttt{python:3.11-slim}) with repository clone, patch application, and test execution---validates functional correctness.

\section{Results}
\label{sec:results}

\subsection{Perfect Separation of Contamination}

Table~\ref{tab:master} shows results for all 9 problems. CCV achieves complete separation between contaminated and genuine reasoning behavior.

\begin{table*}[t]
\centering
\small
\begin{tabular}{lcccccccc}
\toprule
\textbf{Problem} & \textbf{Pre} & \textbf{Actual} & \textbf{Div.} & \textbf{Gold} & \textbf{Time} & \textbf{Reas.} & \textbf{CS} & \textbf{Applied} \\
\midrule
django-11451 & A & Contam. & 0.000 & 1.000 & 4.5s & NO & \textbf{1.000} & 5/5 \\
django-11099 & A & Contam. & 0.000 & 0.911 & 5.5s & NO & \textbf{0.956} & 5/5 \\
astropy-7606 & B & Contam.+Flaw & 0.002 & 0.283 & 7.3s & NO & \textbf{0.641} & 0/5 \\
\midrule
astropy-13236 & A$^\dagger$ & Genuine & 0.451 & 0.297 & 16.8s & FULL & 0.509 & 5/5 \\
sklearn-14894 & C & Genuine & 0.592 & 0.211 & 16.5s & FULL & 0.406 & 2/5 \\
django-10097 & B & Gen.+Flaw & 0.699 & 0.354 & 19.9s & FULL & 0.429 & 1/5 \\
xarray-3151 & C & Genuine & 0.714 & 0.394 & 21.4s & FULL & 0.457 & 0/5 \\
pytest-7571 & C & Genuine & 0.581 & 0.414 & 17.0s & FULL & 0.529 & 0/5 \\
matplotlib-20488 & B & Gen.+Flaw & 0.557 & 0.093 & 19.1s & FULL & 0.368 & 0/5 \\
\bottomrule
\end{tabular}
\caption{CCV results for 9 SWE-bench Verified problems. Div., Gold, Time, Reas., and CS are computed from \emph{lite-mode} trials (issue text only, no repository). Applied = patches successfully applied in separate \emph{full-mode} Docker runs (\texttt{python:3.11-slim}). Rows above the mid-rule show contaminated behavior ($\text{CS} \geq 0.6$); below show genuine reasoning ($\text{CS} < 0.6$). $^\dagger$Pre-labeled as contamination suspect; CCV reclassifies as genuine reasoning.}
\label{tab:master}
\end{table*}

\begin{figure}[t]
\centering
\begin{tikzpicture}
\begin{axis}[
    width=\columnwidth, height=5.5cm,
    xlabel={Code Diversity},
    ylabel={Gold BLEU},
    xlabel style={font=\small},
    ylabel style={font=\small},
    xmin=-0.05, xmax=0.8,
    ymin=-0.05, ymax=1.15,
    grid=both, grid style={gray!20},
    legend style={font=\tiny, at={(0.98,0.98)}, anchor=north east, draw=gray!50, fill=white, fill opacity=0.9},
    tick label style={font=\tiny},
    clip=false,
]
\addplot[only marks, mark=square*, mark size=3pt, red!80!black]
    coordinates {(0.000, 1.000) (0.000, 0.911)};
\addlegendentry{Contaminated}
\addplot[only marks, mark=diamond*, mark size=4pt, orange!80!black]
    coordinates {(0.002, 0.283)};
\addlegendentry{Contam.+Flaw}
\addplot[only marks, mark=*, mark size=2.5pt, blue!70!black]
    coordinates {(0.451, 0.297) (0.592, 0.211) (0.581, 0.414) (0.714, 0.394)};
\addlegendentry{Genuine}
\addplot[only marks, mark=triangle*, mark size=3pt, cyan!70!black]
    coordinates {(0.699, 0.354) (0.557, 0.093)};
\addlegendentry{Gen.+Flaw}
\node[font=\tiny, anchor=south west] at (axis cs:0.02,1.010) {dj-11451};
\node[font=\tiny, anchor=north west] at (axis cs:0.02,0.895) {dj-11099};
\node[font=\tiny, anchor=north] at (axis cs:0.002,0.24) {as-7606};
\node[font=\tiny, anchor=south] at (axis cs:0.451,0.31) {as-13236};
\end{axis}
\end{tikzpicture}
\caption{Code diversity vs.\ gold BLEU for 9 problems. Three clusters emerge: contaminated (top-left, zero diversity, high gold match), contamination-flaw composite (bottom-left, zero diversity, low gold match), and genuine reasoning (right, high diversity).}
\label{fig:scatter}
\end{figure}
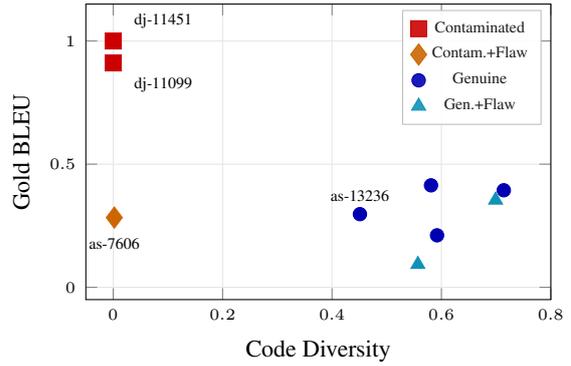

\paragraph{Statistical test.} Mann-Whitney $U = 0$ comparing contaminated ($n=3$; scores 1.000, 0.956, 0.641) versus genuine ($n=6$; scores 0.509, 0.529, 0.457, 0.406, 0.368, 0.429). Exact one-tailed $p = 1/84 \approx 0.012$, effect size $r = 1.0$ (complete separation). Every contaminated score exceeds every genuine score. The minimum gap of 0.112 occurs between the contamination-flaw composite \texttt{astropy-7606} (0.641, the lowest contaminated score) and \texttt{pytest-7571} (0.529, the highest genuine score), with the natural threshold at 0.6. Figure~\ref{fig:cs_distribution} visualizes this separation; Figure~\ref{fig:scatter} shows the underlying diversity--gold proximity structure.

\begin{figure}[t]
\centering
\begin{tikzpicture}
\begin{axis}[
    width=\columnwidth, height=5.5cm,
    xbar, bar width=7pt,
    xlabel={Contamination Score},
    xmin=0, xmax=1.25,
    clip=false,
    ytick={1,2,3,4,5,6,7,8,9},
    yticklabels={matplotlib-20488, sklearn-14894, django-10097, xarray-3151, astropy-13236, pytest-7571, astropy-7606, django-11099, django-11451},
    yticklabel style={font=\tiny},
    xlabel style={font=\small},
    ytick style={draw=none},
    enlarge y limits=0.12,
    nodes near coords,
    nodes near coords style={font=\tiny, anchor=west},
    every node near coord/.append style={xshift=2pt},
]
\addplot[fill=blue!40, draw=blue!60, bar shift=0pt] coordinates {
    (0.368, 1) (0.406, 2) (0.429, 3) (0.457, 4) (0.509, 5) (0.529, 6)
};
\addplot[fill=red!50, draw=red!70, bar shift=0pt] coordinates {
    (0.641, 7) (0.956, 8) (1.000, 9)
};
\draw[dashed, thick, black!60] (axis cs:0.6,0) -- (axis cs:0.6,10)
    node[above, font=\tiny] {threshold};
\end{axis}
\end{tikzpicture}
\caption{Contamination score distribution. Red bars: contaminated behavior ($\text{CS} \geq 0.6$); blue bars: genuine reasoning ($\text{CS} < 0.6$). The dashed line at 0.6 achieves perfect separation with no overlap.}
\label{fig:cs_distribution}
\end{figure}
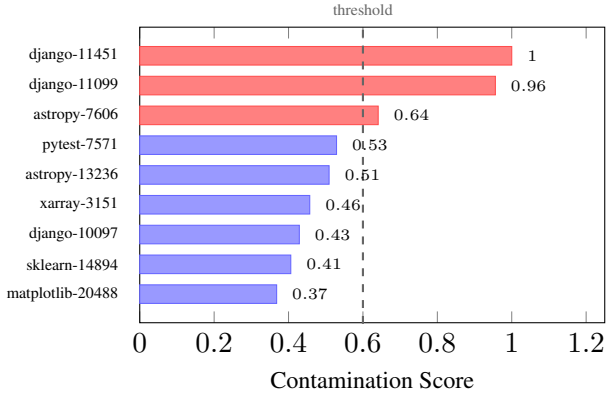

\subsection{Evidence for Binary Contamination}

Contamination scores cluster at extremes: HIGH (0.956--1.000) and LOW (0.368--0.529), with no scores in the 0.55--0.64 range except for the contamination-flaw composite (0.641). Our data is consistent with contamination being a binary phenomenon---models either fully recall a solution or engage in genuine reasoning---though with only $n=3$ contaminated problems (2 HIGH, 1 MEDIUM composite), we cannot rule out intermediate cases at larger scale.

The two confirmed contaminated problems (\texttt{django-11451}, \texttt{django-11099}) exhibit identical profiles: code diversity = 0.0 (all 5 trials byte-identical), gold BLEU $> 0.9$, response time $< 6$s, output tokens $< 260$, and no reasoning process. In full mode, both achieve 100\% patch application and 100\% test resolution---memorized patches are so precise they apply correctly \emph{without the model seeing the repository}.

\subsection{Reasoning as Discriminator}

The reasoning classifier (\S3.4) separates all 15 contaminated-behavior trials (\texttt{NO\_REASONING}) from all 30 genuine-reasoning trials (\texttt{FULL\_REASONING}) with no errors. This is cleaner than code diversity (which has a gray zone for \texttt{astropy-7606} at 0.002) and requires no metric computation---only inspecting the first 100 output tokens.

Response time provides independent corroboration: contaminated problems average 5.77s versus 18.45s for genuine reasoning ($3.2\times$ faster), consistent with memory retrieval versus active computation.

\subsection{False Positive Detection}

\texttt{astropy-13236} was pre-labeled as contamination-suspected based on prior reports of Claude Opus 4.5 reproducing inline comments. However, CCV classifies it as LOW (0.509) with a genuine reasoning profile: diversity 0.451, 5 unique implementation variants, reasoning present, response time 16.8s. Its metrics fall squarely within the control group range (0.368--0.529).

This demonstrates that contamination is conditioned on the \emph{(problem, model, time)} triple---not a fixed property of problems. A problem contaminated for one model version may be clean for another. CCV provides a tool to verify existing contamination labels against actual model behavior, detecting false positives that static methods cannot identify.

\subsection{Test Flaw Detection}
\label{sec:testflaw}

Group B reveals that test-flaw problems are \emph{not} a homogeneous category:

\paragraph{Type B1: Contamination + flaw composite.} \texttt{astropy-7606} shows contamination signals (diversity 0.002, no reasoning, 7.3s response) but produces a non-gold solution (gold BLEU 0.283). The gold patch itself fails in Verified---external evidence of test flaw. The model memorized a \emph{different} solution, possibly from community discussions. This composite nature was discovered through HCCA's independent analysis structure (\S\ref{sec:hcca}).

\paragraph{Type B2: Pure flaw with genuine reasoning.} \texttt{matplotlib-20488} and \texttt{django-10097} show genuine reasoning profiles (diversity 0.557--0.699, reasoning present, 19--20s response) but their diverse, plausible solutions all fail tests. \texttt{matplotlib-20488} is notable: \texttt{files\_overlap}$= 0.0$ between model patches and gold---the model modifies a different file entirely, suggesting the test enforces a specific implementation path.

\section{Discussion}
\label{sec:discussion}

\paragraph{Binary contamination.} Prior work implicitly assumes contamination exists on a spectrum. Our data is consistent with a binary model: scores are either $\geq 0.95$ or $\leq 0.53$, with the sole intermediate case (0.641) being a contamination-flaw composite. If this pattern holds at larger scale, binary classification would suffice for detection---but validation on more contaminated problems is needed to confirm the absence of intermediate cases.

\paragraph{Why reasoning absence works.} When a model has memorized a solution, the answer is immediately available, eliminating the need for analysis. This produces a distinctive output pattern: immediate patch generation with no analytical preamble. While models could theoretically learn to generate ``fake reasoning'' to mask contamination, current models do not exhibit this behavior. Detecting such adversarial adaptation would require analyzing reasoning \emph{quality}, not just presence.

\paragraph{CCV versus existing methods.} Table~\ref{tab:comparison} summarizes key differences. CCV is unique in directly observing the generation process, detecting both contamination and test flaws, and operating in a fully black-box setting without requiring benchmark modifications.

\begin{table}[t]
\centering
\small
\begin{tabular}{lcccc}
\toprule
& \textbf{CAP} & \textbf{SWE-b+} & \textbf{PPL} & \textbf{CCV} \\
\midrule
Approach & Para. & Text & Stat. & \textbf{Gen.} \\
Black-box & \checkmark & \checkmark & $\times$ & \checkmark \\
Observes gen. & $\times$ & $\times$ & $\times$ & \checkmark \\
Test flaws & $\times$ & $\times$ & $\times$ & \checkmark \\
No bench. mod. & $\times$ & \checkmark & \checkmark & \checkmark \\
\bottomrule
\end{tabular}
\caption{CCV vs.\ existing contamination detection. SWE-b+=SWE-bench+, PPL=perplexity-based, Para.=paraphrase, Gen.=generation-based. ``Observes gen.''=directly observes generation process. ``No bench.\ mod.''=requires no benchmark modification.\textsuperscript{$\dagger$}CAP requires generating paraphrased versions of benchmark problems, which we classify as benchmark modification.}
\label{tab:comparison}
\end{table}

\paragraph{HCCA as methodology.} The hierarchical architecture prevented confirmation bias: the Group~C analyst independently identified \texttt{astropy-7606}'s composite nature (CS=0.641) that the Group~B analyst had classified as pure test flaw. This cross-validation discovery became a key finding. While CCV can function without HCCA, the hierarchical structure adds analytical rigor valuable when results carry high stakes (e.g., benchmark curation decisions).

\paragraph{The primacy of information restriction.} Three experiments converge on a single principle. CCR \cite{ccr2026} showed that session-isolated review (where the reviewer cannot see the production context) improves quality by 4.0 F1 points. D-CCR \cite{song2026dccr} showed that repeating the same role with access to prior conclusions \emph{degrades} quality (F1: 0.376 $\to$ 0.303, $p < 0.001$). Our pilot (\S\ref{sec:pilot}) showed that structural division of labor (Worker/Verifier/Director) with shared conclusions produces \emph{zero} filtering---100\% sycophantic confirmation. The common thread is information restriction: when each analytical stage makes judgments without seeing prior conclusions, verification improves; when prior conclusions are visible, regardless of role differentiation or pipeline complexity, sycophancy and anchoring eliminate any benefit. This suggests that information restriction is not merely a useful technique but the \emph{necessary condition} for effective LLM-based verification.

\paragraph{From repetition to structure: a research program.} This work is part of a three-study program on context separation in LLM verification. CCR \cite{ccr2026} established that separating production and review contexts improves quality. D-CCR \cite{song2026dccr} showed that repeating the same reviewer role across multiple rounds \emph{degrades} quality---the optimal number of review rounds is one. HCCA's CCV analysis demonstrates that \emph{structural division of labor} across different analytical roles \emph{with information restriction} succeeds, while the pilot experiment (\S\ref{sec:pilot}) confirms that structural roles \emph{without information restriction} fail. The unifying principle is that what matters is not the amount of verification or the diversity of roles, but whether each stage judges \emph{independently}---paralleling CCV's own detection mechanism, which relies on solution diversity produced by session isolation.

\paragraph{Beyond benchmark verification.} While this study applies HCCA to contamination detection, the architecture generalizes to any analysis requiring multiple independent perspectives. The core design---specialized roles, information restriction, hierarchical integration---could benefit collaborative code review, literature synthesis, or statistical verification. Systematic evaluation of such applications remains future work.

\section{Limitations}
\label{sec:limitations}

\paragraph{Sample size.} Nine problems provide statistical significance ($p \approx 0.012$) with maximal effect size ($r = 1.0$), but limited generalizability. The binary contamination finding and 0.6 threshold require validation on larger samples. We have an ongoing 30-problem extension using the same methodology, and release all code and data to enable independent replication at any scale.

\paragraph{Single model.} We evaluate only Claude Opus 4.6. Cross-model validation (GPT, Gemini, open-source models) would strengthen claims about CCV's generality. Our false positive finding (\texttt{astropy-13236}: contaminated in Opus 4.5, clean in Opus 4.6) already suggests model-version sensitivity.

\paragraph{Python only.} SWE-bench is Python-exclusive. CCV's applicability to other languages and domains (mathematics, science QA) is untested, though the core principle (diversity from repeated independent solving) is language-agnostic.

\paragraph{Reasoning classifier.} Our keyword-based classifier achieves 100\% accuracy on this dataset but is evaluated on the same data used for detection (circular validation). The behavioral independence of reasoning signals from diversity metrics partially mitigates this concern (\S3.4), but validation on held-out problems is essential. The classifier could also be defeated by models trained to generate synthetic reasoning traces before memorized outputs.

\paragraph{Formula limitations.} The test flaw score (Eq.~\ref{eq:testflaw}) degenerates when code diversity approaches 0. Our correction (\S3.3) addresses the known case (memorized non-gold solutions) but may not cover all edge cases.

\paragraph{HCCA overhead.} The hierarchical architecture requires multiple independent analysis sessions, increasing computational cost. For straightforward cases where contamination signals are unambiguous, a single-analyst CCV analysis may suffice. HCCA's value is clearest in ambiguous cases (e.g., contamination-flaw composites) where independent perspectives prevent premature conclusions.

\section{Ethical Considerations}

CCV is designed for benchmark integrity verification, but we acknowledge potential misuse. First, knowledge of CCV's detection mechanism could enable adversarial training to produce ``fake diversity''---varying surface-level code while preserving memorized logic. We view this as unlikely in practice (diversity at the AST level is harder to fake than text-level variation) but worth monitoring. Second, publicly labeling specific benchmark problems as contaminated could unfairly penalize model providers. We focus on problem-level diagnosis rather than model blame, and note that contamination often reflects training data curation rather than intentional gaming. We release our methodology and data to enable transparent, reproducible verification by the community.

\section{Conclusion}

We introduce Cross-Context Verification (CCV), a black-box contamination detection method, and the Hierarchical Cross-Context Architecture (HCCA), a multi-agent analysis framework that prevents confirmation bias through intentional information restriction. On 9 SWE-bench Verified problems, CCV achieves perfect separation ($U=0$, $p \approx 0.012$, $r = 1.0$), while HCCA's independent analysis structure discovers contamination-flaw composites that single-analyst approaches miss.

Our findings---binary contamination, reasoning as discriminator, 33\% false positive detection---make CCV a label verification tool, not just a detection method. Together with CCR's session isolation \cite{ccr2026}, D-CCR's negative result on repetition \cite{song2026dccr}, and HCCA's pilot showing that structural roles without information restriction produce zero benefit (\S\ref{sec:pilot}), a consistent picture emerges: \emph{information restriction is the necessary condition} for effective LLM-based verification. Structural complexity, role differentiation, and additional rounds all fail without it. Future work should explore methods that maintain information restriction while improving verification accuracy---the key constraint is what each stage is allowed to \emph{see}, not what it is asked to \emph{do}. We release all code, data, and analysis to support reproducibility and adoption.


\bibliography{references}

\appendix

\section{Full Trial Data}
\label{app:trials}

Table~\ref{tab:full_trials} reports per-problem statistics across all 45 trials (9 problems $\times$ 5 trials).

\begin{table*}[t]
\centering
\small
\begin{tabular}{lccccccccc}
\toprule
\textbf{Problem} & \textbf{Grp} & \textbf{Trials} & \textbf{Diversity} & \textbf{Gold $\bar{g}$} & \textbf{Gold $\sigma_g$} & \textbf{Unique} & \textbf{Time (s)} & \textbf{Tokens} & \textbf{Reas.} \\
\midrule
django-11451 & A & 5 & 0.000 & 1.000 & 0.000 & 1 & 4.5 & 154 & NO \\
django-11099 & A & 5 & 0.000 & 0.911 & 0.000 & 1 & 5.5 & 256 & NO \\
astropy-13236 & A & 5 & 0.451 & 0.297 & 0.023 & 5 & 16.8 & 1156 & FULL \\
astropy-7606 & B & 5 & 0.002 & 0.283 & 0.000 & 1 & 7.3 & 473 & NO \\
matplotlib-20488 & B & 5 & 0.557 & 0.093 & 0.059 & 5 & 19.1 & 1246 & FULL \\
django-10097 & B & 5 & 0.699 & 0.354 & 0.194 & 4 & 19.9 & 1320 & FULL \\
sklearn-14894 & C & 5 & 0.592 & 0.211 & 0.112 & 4 & 16.5 & 1164 & FULL \\
pytest-7571 & C & 5 & 0.581 & 0.414 & 0.018 & 5 & 17.0 & 1084 & FULL \\
xarray-3151 & C & 5 & 0.714 & 0.394 & 0.129 & 5 & 21.4 & 1424 & FULL \\
\bottomrule
\end{tabular}
\caption{Per-problem statistics across all 45 trials (9 problems $\times$ 5 trials each). Diversity and Gold $\bar{g}$ are computed via Eq.~\ref{eq:contamination}; Gold $\sigma_g$ is the standard deviation of per-trial gold BLEU scores. Unique = number of distinct implementation approaches (manual inspection). Reas.\ = reasoning classifier output (NO = \texttt{NO\_REASONING}, FULL = \texttt{FULL\_REASONING}).}
\label{tab:full_trials}
\end{table*}

\paragraph{Full-mode Docker results.} Table~\ref{tab:docker} reports patch application and test resolution rates from Docker runs.

\begin{table}[t]
\centering
\small
\begin{tabular}{lccc}
\toprule
\textbf{Problem} & \textbf{Reas.} & \textbf{Applied} & \textbf{Resolved} \\
\midrule
django-11451 & NO & 5/5 & 5/5 \\
django-11099 & NO & 5/5 & 5/5 \\
astropy-13236 & FULL & 5/5 & 0/5 \\
astropy-7606 & NO & 0/5 & 0/5 \\
matplotlib-20488 & FULL & 0/5 & 0/5 \\
django-10097 & FULL & 1/5 & 0/5 \\
sklearn-14894 & FULL & 2/5 & 0/5 \\
pytest-7571 & FULL & 0/5 & 0/5 \\
xarray-3151 & FULL & 0/5 & 0/5 \\
\bottomrule
\end{tabular}
\caption{Full-mode Docker test results. Only pure contamination cases (\texttt{django-11451}, \texttt{django-11099}) achieved test resolution (10/10 trials). The contamination-flaw composite (\texttt{astropy-7606}, also \texttt{NO\_REASONING}) and all genuine-reasoning problems achieved 0/35 resolution.}
\label{tab:docker}
\end{table}

\section{Scoring Formula Derivation}
\label{app:scoring}

\paragraph{Code diversity (Eq.~1).} Pairwise similarity combines three complementary signals:
\begin{itemize}
\itemsep0em
\item \textbf{AST structural similarity} (weight 0.4): Uses Zhang-Shasha tree edit distance on Python AST nodes, normalized by tree size. Captures structural equivalence regardless of variable naming or formatting.
\item \textbf{Line-level BLEU} (weight 0.3): Standard BLEU score on patch lines, capturing surface-level similarity including comments and formatting.
\item \textbf{Normalized edit distance} (weight 0.3): Character-level Levenshtein distance normalized by maximum length. Captures fine-grained textual differences.
\end{itemize}
AST receives the highest weight because structurally identical code with cosmetic differences (variable names, whitespace) should be classified as the same solution.

\paragraph{Contamination score (Eq.~2).} Gold proximity ($\bar{g}$, weight 0.5) is the strongest contamination indicator: verbatim reproduction of the gold patch is direct evidence of memorization. Code diversity ($1 - d$, weight 0.3) captures solution space exploration independent of gold match. Gold proximity variance ($1 - \sigma_g$, weight 0.2) detects recall consistency; memorized solutions show zero variance.

\paragraph{Test flaw score (Eq.~3).} Functional equivalence ratio (FER, weight 0.4) measures whether diverse solutions converge to the same test outcomes. Diverse-but-failing ratio (DBF, weight 0.4) captures how many plausibly correct solutions the test rejects. Code diversity ($d$, weight 0.2) provides context: high diversity with high DBF strongly suggests test issues.

\paragraph{Memory-of-non-gold correction (\S3.3).} When diversity $< 0.05 \wedge \bar{g} < 0.5$, the model has memorized a non-gold solution. The standard diversity term (0.2$d$) contributes $\approx 0$ despite strong evidence of test flaw (a memorized solution that differs from gold). Replacing with $0.2(1 - \bar{g})$ ensures the score reflects the gold--memorized divergence.

\section{HCCA Implementation Details}
\label{app:hcca}

In our implementation, HCCA uses 6 terminals across 4 layers: T0 (L0 Designer), T5 (L1 Executor), T2/T3/T4 (L2 Analysts for Groups A/B/C respectively), and T1 (L3 Integrator). Information flows T5 $\to$ T2/T3/T4 $\to$ T1, with no lateral communication at Layer~2.

Each terminal operates as an independent Claude Code session with its own context window. Terminal state is persisted via markdown files, enabling session recovery without context contamination. T4's cross-validation of T2/T3 results occurs after completing its own Group~C analysis, using only L1 data (not L2 conclusions from T2/T3). This ensures that cross-validation is genuinely independent rather than anchored on prior analytical conclusions. See Appendix~\ref{app:hcca} for full implementation details.

\section{Problem Selection Rationale}
\label{app:selection}

\paragraph{Group A (contamination suspected).} Problems were selected based on documented evidence of verbatim or near-verbatim reproduction by frontier models:
\begin{itemize}
\itemsep0em
\item \texttt{django-11451}: GPT-5.2 reproduced exact gold patch including comments
\item \texttt{django-11099}: Gemini 3 Flash produced byte-identical solution
\item \texttt{astropy-13236}: Claude Opus 4.5 included inline citations suggesting training data recall
\end{itemize}

\paragraph{Group B (test flaws).} Problems were selected based on external evidence of test-side issues:
\begin{itemize}
\itemsep0em
\item \texttt{astropy-7606}: Gold patch fails when applied in SWE-bench Verified harness
\item \texttt{matplotlib-20488}: SSL certificate and C extension build issues cause flaky failures
\item \texttt{django-10097}: Tests enforce specific template path handling not specified in the issue
\end{itemize}

\paragraph{Group C (control).} Selected for: (1)~no prior contamination or test flaw evidence, (2)~sub-50\% solve rates on public leaderboards, (3)~repository diversity (scikit-learn, pytest, xarray---none duplicating Group~A/B repos).

\paragraph{Excluded problems (9).}
\begin{itemize}
\itemsep0em
\item \texttt{psf/requests}: 4 problems excluded (repository overrepresentation)
\item \texttt{django-10914}, \texttt{django-10924}: Ambiguous contamination signals
\item \texttt{matplotlib-23913}: Environmental dependency issues
\item \texttt{seaborn-2848}: Limited community evidence
\item \texttt{sympy-13895}: Symbolic math domain (different failure modes)
\end{itemize}

\end{document}